\documentclass[11pt]{article}

\usepackage[final]{acl}
\usepackage{times}
\usepackage{latexsym}
\usepackage[T1]{fontenc}
\usepackage[utf8]{inputenc}
\usepackage{microtype}
\usepackage{graphicx}
\usepackage{float}
\usepackage{booktabs}
\usepackage{multirow}
\usepackage{amsmath}

\title{SZED at ArGuard Shared Tasks: Multi-Task Label-Attention and \\ Cross-Modal Fusion for Arabic Hateful Meme Detection}

\author{
Selma Khelili \\
ENSIA, Algeria \\
\texttt{selma.khelili@ensia.edu.dz} \\
\And
Zohra Abdeli \\
ENSIA, Algeria \\
\texttt{zohra.abdeli@ensia.edu.dz} \\
}

\begin{document}
\maketitle

\begin{abstract}
We present SZED's submission to ArGuard 2026 Track~A: Arabic hateful meme detection.
The task has two parts: binary classification (A1) and fine-grained multi-label
tagging (A2).
For A1, we fuse MARBERTv2 and CLIP ViT-L/14 embeddings, comparing concatenation,
self-attention, and ensemble heads.
For A2, we use a multi-task label-attention architecture with per-class query vectors
and an auxiliary head for cascade calibration.
We reach macro-F1 = 0.7597 on A1 and 0.3142 on A2 (averaged over 10 active classes).
We report reproducibility across runs, per-class error patterns, and the effect of
cascade calibration on the long-tailed A2 label space.
\end{abstract}

\section{Introduction}

Detecting hateful content in Arabic multimodal memes is harder than detecting
text-only hate speech: it requires jointly interpreting visual cues, embedded OCR
text, dialectal variation, sarcasm, and culturally specific references
\cite{alam-etal-2026-arguard}.
ArGuard 2026 Track~A splits this problem into two stages.
Subtask~A1 asks whether a meme is hateful at all, while Subtask~A2 requires
assigning zero or more fine-grained harm categories (e.g., \textit{Mocking},
\textit{Incitement}, \textit{Dehumanization}) to memes regardless of their binary
status, over a label space that partially overlaps with the AHA-Memes benchmark
\cite{kmainasi2026ahamemes}.

We treat A1 and A2 as architecturally distinct problems.
A1 uses a standard late-fusion design, and most of our effort there goes into
comparing simple fusion strategies.
For A2, a single pooled fusion vector seemed like a poor fit for ten simultaneous,
structurally-linked, and heavily imbalanced binary decisions, so we instead use
per-class label-attention together with an auxiliary binary signal for cascade
calibration.
As an ablation on A1, we also test whether a more expressive cross-modal attention
mechanism and an explicit image-text incongruity feature improve on the simpler
ensemble.
We found that they do not, and Section~\ref{sec:ablation} discusses why this
negative result is still informative.
Our code is publicly
available.\footnote{\url{https://github.com/SelmaKhelili/ARGUARD-2026-TASK-A}}

\section{Related Work}

\paragraph{Arabic and multilingual hateful meme detection.}
Hateful meme detection has been widely studied for English-language content \cite{tan-bansal-2019-lxmert,lu2019vilbert}. However, Arabic remains relatively under-resourced. The AHA-Memes benchmark \cite{kmainasi2026ahamemes} is closely related to our work. It contains 5K Arabic memes with fine-grained multi-label annotations. Its taxonomy also overlaps substantially with the A2 label space used in ArGuard.

The results reported on AHA-Memes highlight several important challenges. Text provides the strongest unimodal signal for Arabic memes, as text-only fine-tuned models outperform image-only models. Late fusion further improves binary hateful recall compared with text-only models. However, fine-grained hate-type classification remains difficult across model families. The best reported macro-F1 is 0.340, achieved by a large vision-language model \cite{kmainasi2026ahamemes}. These results provide a useful reference for the difficulty of the A2 task. However, the AHA-Memes benchmark uses a different data collection process and train/test split from ArGuard.

\paragraph{Multimodal fusion for meme understanding.}
A common approach to multimodal classification is late fusion. In this setting, modality-specific representations, such as CLS or pooled embeddings, are combined and passed to a shared classifier \cite{tan-bansal-2019-lxmert,lu2019vilbert}. This approach can work well for binary classification. However, it can be limiting for fine-grained multi-label tasks with strong class imbalance.

A shared representation requires the model to make several label decisions from the same bottleneck vector. Label-attention methods provide an alternative. Each class is represented by a learned query that attends to the full sequence of text tokens and image patches \cite{ridnik2021asymmetric}. This allows each label to focus on the parts of the multimodal input that are most relevant to it. We use this design for the A2 task.

\paragraph{Research gap.}
Previous Arabic meme studies, including AHA-Memes baselines, generally use a shared fusion head for both binary and fine-grained classification. In ArGuard, we observe strong co-occurrence patterns between the binary hateful label and the fine-grained hate labels (Section~\ref{sec:data}). This structure provides useful information that a shared classification head does not explicitly model.

Our approach addresses this gap in two ways. First, we use cascade calibration to explicitly incorporate the binary prediction into the fine-grained decision process. Second, we use per-class label attention instead of a single pooled representation for multi-label prediction. Together, these choices allow the model to better exploit the structure of the A2 label space.

\section{Data and Preprocessing}
\label{sec:data}

\begin{table}[H]
\centering
\small
\begin{tabular}{lrrr}
\toprule
Split & \# examples & \# Hateful & \% Hateful \\
\midrule
Train                  & 3{,}500 & 1{,}324 & 37.8 \\
Dev                    &     500 &     189 & 37.8 \\
Dev-test               &     500 &     189 & 37.8 \\
Test (official)        &     500 &     148 & 29.6 \\
\bottomrule
\end{tabular}
\caption{ArGuard Task~1 split sizes and class balance.
Train, dev, and dev-test share an identical 37.8\% Hateful rate; the official
test split is more balanced at 29.6\% Hateful.
The dev-test split is used in the backbone-comparison experiments
(Appendix~\ref{sec:appendix}).}
\label{tab:data}
\end{table}
We use the official ArGuard Task~1 data released via the Hugging Face Hub
(\texttt{QCRI/ArGuard-Task1}): Arabic memes paired with OCR-extracted text and
binary/fine-grained labels.
A1 uses two active classes (\textit{Hateful}, \textit{Not Hateful}) and A2 uses
ten (\textit{Contempt, Dehumanization, Exclusion, Humor, Incitement, Inferiority,
Mocking, Other, Sarcasm, Slurs}).
Five further categories in the task schema
(\textit{Extremism, Historical, Insults, Stereotyping, Threat}) have zero support
in our training split and are excluded entirely from scoring by the official
scorer;\footnote{Verified against \texttt{scorer/scorer.py} and
\texttt{baselines/labels.py} in the official ArGuard-2026-tasks repository, which
the organizers confirm is the same script CodaBench runs server-side against the
held-back gold.
All A2 macro-F1 figures in this paper are therefore averaged over the
10~active classes only, not all~15.}
we never predict them.
Split sizes are summarized in Table~\ref{tab:data}.

The ten A2 labels are heavily imbalanced: \textit{Sarcasm} has 934 training
instances and \textit{Humor} 863, while \textit{Exclusion} has just 10.
We also observed a near-deterministic structural link between A2 labels and the
binary label: \textit{Humor} and \textit{Sarcasm} never co-occur with
\textit{Hateful} in training, and most remaining classes co-occur with it in
100\% of instances.
\textit{Other} is the exception ($\approx$4.5\% Hateful co-occurrence).
This pattern motivates the cascade-calibration design described in
Section~\ref{sec:a2}.

Text is tokenized with the MARBERTv2 tokenizer (max length 128), and images are
processed with the corresponding CLIP image processor at 224px.
Beyond what is described above, we use
no external training data, synthetic data, or data
augmentation.

\section{System Architecture and Training}

\begin{figure}[t]
\centering
\includegraphics[width=\columnwidth]{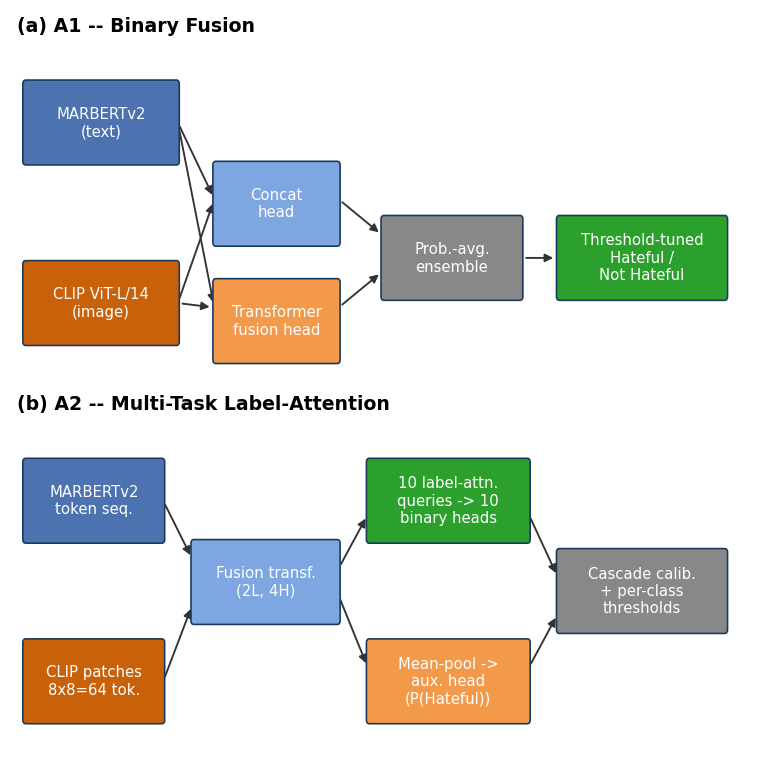}
\caption{System overview. (a) A1: shared encoders feed two fusion heads combined
via a probability-averaged ensemble. (b) A2: full token/patch sequences feed a
fusion transformer, read by ten label-attention queries and, separately, an
auxiliary binary head used for cascade calibration.}
\label{fig:architecture}
\end{figure}

Both subtasks share the same pretrained encoder pair
(Figure~\ref{fig:architecture}): MARBERTv2 \cite{abdul-mageed-etal-2021-arbert}
for Arabic text (hidden size 768) and CLIP ViT-Large/14 \cite{radford2021clip}
for images (hidden size 1024).
All models are implemented in PyTorch using the HuggingFace transformers library.
To fit training within T4/L4 GPU memory, we freeze all layers except the last two
transformer blocks of each encoder, leaving 8.7--8.9\% of parameters trainable
(A1: 40.4M / 466.5M; A2: 41.8M / 467.8M).
The shared fusion space has dimension 256; all MLP classifiers use hidden width 256.
All models are trained with AdamW \cite{loshchilov2019decoupled}
(lr~=~2e\mbox{-}5, weight decay~=~1e\mbox{-}4) and linear warmup (6\%)
followed by decay, batch size~8, gradient accumulation~2 (effective batch~16),
mixed precision, and gradient clipping at norm~1.0. All experiments use a single fixed seed (42).

\subsection{Subtask A1: Binary Fusion}

On top of the shared encoders we compare two fusion heads
(Figure~\ref{fig:architecture}a).
The \emph{concatenation head} concatenates the pooled text [CLS] and CLIP
embeddings and passes them through a small MLP classifier
(\texttt{Linear(1792$\to$256)+ ReLU + Dropout $\to$ Linear(256$\to$2); 460K params}).
The \emph{transformer-fusion head} projects both embeddings to the shared 256-dim
space with a learned modality-type embedding, applies a two-layer, four-head
self-attention encoder (FFN width 512; 1.51M params), and then mean-pools before
classifying.
Note that this ``transformer fusion'' applies self-attention over two pooled
vectors, not token/patch-level cross-attention;
Section~\ref{sec:ablation} reports a variant that uses the latter.
Both heads train for 6~epochs with class-weighted cross-entropy
(weights $\approx$0.54/1.46 for the 62/38 class imbalance).
After training, we grid-search a decision threshold (37 points, 0.05-0.95) for
the concat head, the transformer head, and their probability-averaged ensemble,
and select whichever variant/threshold combination maximizes dev macro-F1.

\subsection{Subtask A2: Multi-Task Label-Attention}
\label{sec:a2}

A2 uses a distinct architecture (Figure~\ref{fig:architecture}b), motivated by
the two properties noted in Section~\ref{sec:data}: imbalance across ten
simultaneous decisions and a strong structural link to the binary label.
Rather than pooling each modality into a single vector, we 
MARBERTv2's full
token sequence and CLIP's patch grid (spatially downsampled 2$\times$2,
16$\times$16~$\to$~8$\times$8~=~64 tokens).
We project both to a shared 256-dim space and fuse them with a two-layer,
four-head transformer encoder (FFN width 512).
Each of the ten classes gets its own learned query vector (10$\times$256
parameter table), which attends over the fused sequence through multi-head
label-attention and feeds an independent linear classifier.
This avoids the single-bottleneck failure mode of pooled multi-label
classification.
A mean-pooled summary of the fused sequence also feeds a small auxiliary binary (Hateful/Not Hateful) head (\texttt{Linear(256$\to$128)+ReLU+ Linear(128$\to$2); 33K params}).
The two heads are trained jointly via
$\mathcal{L} = \mathcal{L}_{\text{ASL}}(\text{fg}) + 0.3 \cdot
\mathcal{L}_{\text{CE}}(\text{aux})$,
where $\mathcal{L}_{\text{ASL}}$ is Asymmetric Loss \cite{ridnik2021asymmetric}
($\gamma_{\text{neg}}$=4.0, $\gamma_{\text{pos}}$=1.0, clip=0.05).
We chose ASL over plain weighted BCE for its better fit to long-tailed multi-label imbalance. We train for 10 epochs. Dev macro-F1 at a global 0.5 threshold reaches its peak at epoch 4 and then falls even as training loss continues to decrease, so we keep the epoch-4 checkpoint.

\textbf{Cascade calibration.}
At inference, we exploit the near-deterministic label hierarchy directly.
Hateful-subtype probabilities are scaled by the auxiliary head's P(Hateful),
non-hateful-only subtypes by P(Not~Hateful), and \textit{Other} is left unscaled.
We then tune a per-class decision threshold on these calibrated probabilities
(grid search, 19 points) rather than using one global threshold, since rare
classes (e.g., \textit{Sarcasm}, best threshold 0.10) and common ones
(e.g., \textit{Other}, best threshold 0.55) need different operating points.
Thresholds are tuned on the 500-example dev set, which carries an overfitting risk
for the rarest classes; we flag this in Limitations.

\section{Experimental Results}

\subsection{Subtask A1}

\begin{table}[t]
\centering
\small
\begin{tabular}{lrr}
\toprule
Run & Macro-F1 & Accuracy \\
\midrule
Dev (leaderboard)      & 0.7381 & 0.7620 \\
Test, run 1            & 0.7556 & 0.7880 \\
Test, run 2 (official) & \textbf{0.7597} & 0.7940 \\
\bottomrule
\end{tabular}
\caption{A1 results across independent runs of the identical architecture and
hyperparameters. Test, run~2 is our official leaderboard submission.}
\label{tab:a1-results}
\end{table}

Table~\ref{tab:a1-results} shows our primary A1 results.
We did not persist trained weights between the development and test phases, so
each row reflects an independently trained run of the same architecture and hyperparameters.
Across our two test-phase reruns, macro-F1 ranges from 0.7556 to 0.7597
(F1-Hateful 0.6590-0.6688; F1-Not-Hateful 0.8446-0.8505), suggesting the base
fusion system is reasonably stable under CUDA non-determinism.
One thing worth flagging for reproducibility is that variant selection is less
stable: identical-seed reruns picked \emph{different} best variants at the
threshold-tuning stage, even though all scored within 0.5~percentage points of each other on dev (an absolute macro-F1 difference of 0.005).
The official run (run~2) reached F1-Hateful = 0.669, F1-Not-Hateful = 0.850, and
accuracy = 0.794, corresponding to a confusion matrix of
TN=293, FP=59, FN=44, TP=104 on the 500-example test set:
the system missed 44 of 148 hateful memes (false negatives) while
over-flagging 59 of 352 non-hateful memes (false positives).

We also tested our A1 methodology on three other Arabic text-model variants;
see Appendix~\ref{sec:appendix} for the full comparison.

\subsection{Ablation: Cross-Modal Incongruity}
\label{sec:ablation}

\begin{table}[t]
\centering
\small
\begin{tabular}{lrrrr}
\toprule
Variant & MF1 & Acc. & F1$_H$ & F1$_{NH}$ \\
\midrule
Ensemble    & 0.7340 & 0.754 & 0.6612 & 0.8069 \\
Transformer & 0.7297 & 0.750 & 0.6556 & 0.8038 \\
Concat      & 0.7285 & 0.746 & 0.6595 & 0.7974 \\
Cross-modal & 0.6878 & 0.704 & 0.6166 & 0.7590 \\
\bottomrule
\end{tabular}
\caption{A1 ablation on dev set (one run per variant, seed 42; see
Section~\ref{sec:ablation} for a caveat on single-run comparisons).
MF1 = macro-F1; F1$_H$ / F1$_{NH}$ = per-class F1.
Best threshold per variant applied.
The incongruity signal carries no separating power: mean CLIP cosine similarity
is 0.236 for Hateful and 0.238 for Not Hateful on training data.}
\label{tab:a1-ablation}
\end{table}

Independent of the primary system, we trained a third A1 variant with genuine
token/patch-level bidirectional cross-attention, where text tokens attend over
CLIP patches and vice versa, following a two-layer LXMERT/ViLBERT-style design
\cite{tan-bansal-2019-lxmert,lu2019vilbert}.
We augmented this with a frozen cosine-similarity feature between multilingual-CLIP
text and image embeddings as an explicit image-text incongruity signal.
Table~\ref{tab:a1-ablation} reports all four variants: the cross-modal variant
underperforms the simpler ensemble on dev (macro-F1 0.6878 vs.\ 0.7340), and the
gap between the best and worst variant (0.0462 macro-F1) is clear directly from
the table without needing a chart.
A sanity check explains the gap: the mean incongruity score is nearly identical
between \textit{Hateful} (0.236) and \textit{Not~Hateful} (0.238) memes on
training data, so the auxiliary signal carries no separating information on this
dataset.
We interpret this as a negative result: meme hatefulness here is driven
more by caption semantics and iconography than by caption-image mismatch.
Because this run was trained independently, its numbers differ slightly from
Table~\ref{tab:a1-results} (consistent with run-to-run variance noted in
Section~5.1); the ablation rests on one run per variant.
Given the run-to-run variance we observe elsewhere for this system
(Section~5.1), the 0.0462 macro-F1 gap between the cross-modal and ensemble
variants should be read as suggestive rather than statistically confirmed.
We treat the cosine-similarity sanity check (0.236 vs.\ 0.238) as the primary
evidence for the negative result, since unlike the F1 comparison, it does
not depend on a single training run.

\subsection{Subtask A2}

\begin{table}[t]
\centering
\small
\begin{tabular}{lrrr}
\toprule
Metric & Dev\textsuperscript{a} & Test\textsuperscript{a} & Test\textsuperscript{b} \\
\midrule
Macro-F1 (raw @0.5)   & 0.3276 & --     & 0.2969 \\
Macro-F1 (calibrated) & 0.3450 & 0.3142 & 0.3092 \\
Micro-F1               & --     & 0.4557 & 0.4395 \\
Hamming loss            & --     & 0.1992 & 0.2112 \\
Subset accuracy         & --     & 0.1460 & 0.1340 \\
\bottomrule
\end{tabular}
\caption{A2 results across two runs of the identical architecture.
\textsuperscript{a}Submitted/official run (dev and official leaderboard test
scores). \textsuperscript{b}An independent rerun, evaluated on test only.
Raw: sigmoid outputs at a global 0.5 threshold, no cascade calibration.
Calibrated: cascade-scaled probabilities with per-class thresholds.
All macro-F1 figures are averaged over the 10 active classes
(Section~\ref{sec:data}). The calibrated test macro-F1 gap between the rerun
(0.3092) and the official submission (0.3142) reflects run-to-run variance,
consistent with the effect discussed for A1 in Section~4.1.}
\label{tab:a2-results}
\end{table}
Table~\ref{tab:a2-results} reports raw (uncalibrated) scores alongside
calibrated scores on both dev and test, allowing a direct comparison on each
split.
Cascade calibration and per-class threshold tuning improve macro-F1 by
+0.0123~absolute on test (0.2969~$\to$~0.3092 in this run) and by
+0.0174~absolute on dev (0.3276~$\to$~0.3450 in the submitted run),
confirming that exploiting the binary-label structure at inference time is
beneficial on both splits.

Across three independent test-phase reruns of the same architecture, official
macro-F1 averages $0.3125 \pm 0.0018$ (0.3107, 0.3127, 0.3142), a tighter spread
than observed for A1, suggesting that the label-attention architecture is
comparatively stable.
The last rerun (0.3142) is our official leaderboard submission;
Figure~\ref{fig:a2-perclass} presents its per-class breakdown.

We also tested our A2 methodology on the same three alternative Arabic models used for A1; see Appendix~\ref{sec:appendix} for the full comparison.

\section{Error Analysis}

\begin{figure}[t]
\centering
\includegraphics[width=\columnwidth]{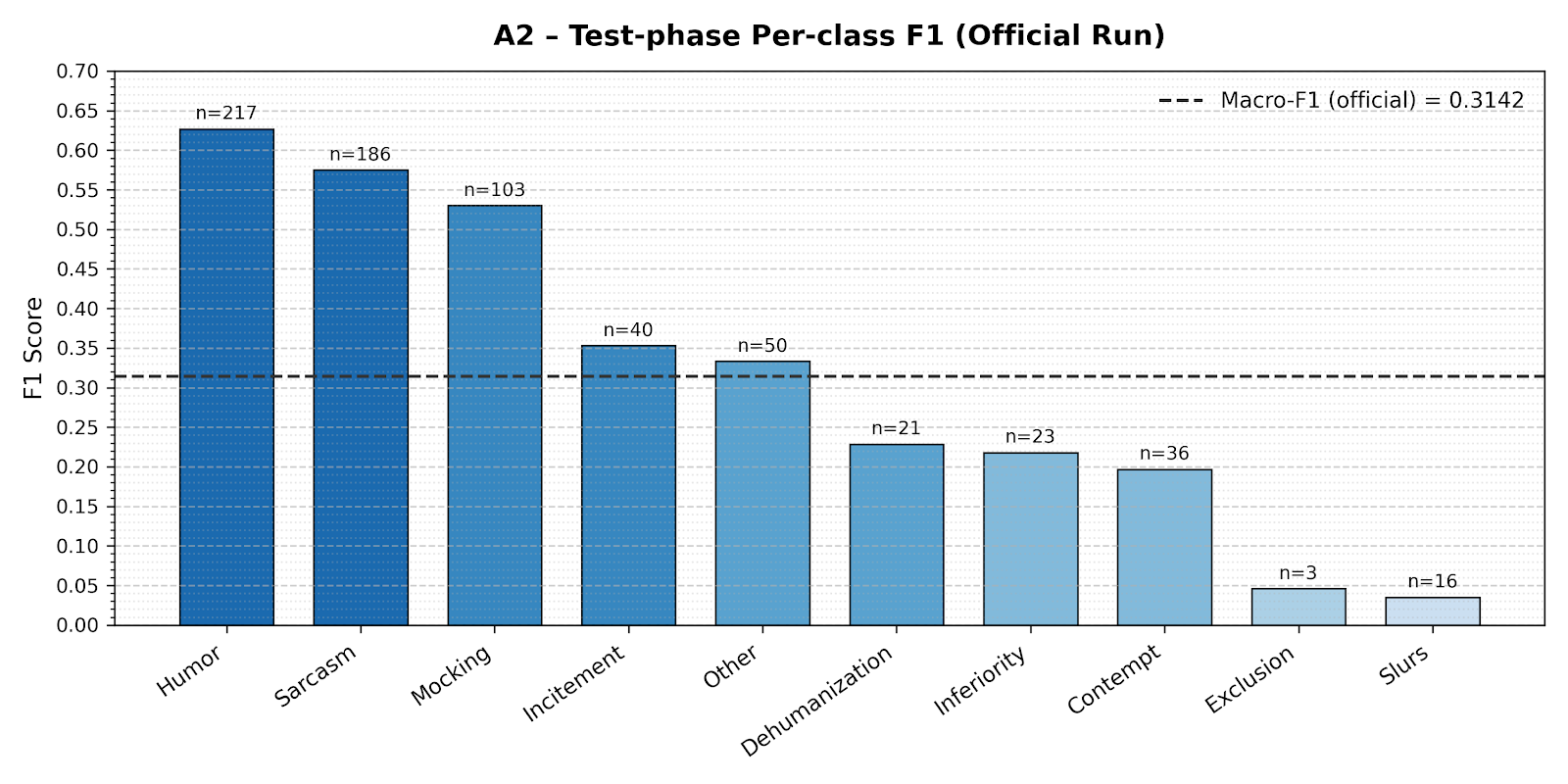}
\caption{Official test-set per-class F1 for A2
(macro-F1 = 0.3142 over 10 active classes), with support (n), sorted descending.}
\label{fig:a2-perclass}
\end{figure}

Per-class F1 on the official A2 test set (Figure~\ref{fig:a2-perclass}) tracks
class support closely.
High-support classes perform reasonably (\textit{Humor}, n=217, F1=0.627;
\textit{Sarcasm}, n=186, F1=0.575; \textit{Mocking}, n=103, F1=0.530), while
low-support classes collapse (\textit{Slurs}, n=16, F1=0.035;
\textit{Exclusion}, n=3, F1=0.046).
We would therefore not interpret these results as evidence that the model is inherently unable to learn these categories: with only 3-16 test instances per class, F1 scores are highly sensitive to sampling variability, and reflect constraints of the evaluation methodology as much as model limitations.

A second pattern is semantic overlap between neighboring categories.
\textit{Mocking}, \textit{Sarcasm}, and \textit{Humor} share surface-level
irony markers and exaggeration cues that are not always distinguishable from
OCR text and image content alone; Arabic sarcasm in particular tends to rely on
discourse-level and cultural context that a token/patch-level model cannot access.
\textit{Slurs} and \textit{Contempt} face dialectal variation and euphemistic
substitution: an offensive term may be transliterated, partially censored, or
expressed in a regional dialect outside MARBERTv2's pretraining distribution.

For A1, the test results are slightly higher than the development result, with macro-F1 increasing from 0.7381 on the development set to 0.7556 and 0.7597 across the two test runs. This does not indicate a clear generalization problem. However, since the development and test models were trained independently, these results should not be interpreted as direct evidence that overfitting is absent.
The per-class results provide more insight than the overall macro-F1. Across the reported runs, F1 for the Not-Hateful class consistently exceeds F1 for the Hateful class by approximately 15-19 points. This indicates that the model has greater difficulty identifying hateful memes.
The official test confusion matrix further illustrates this pattern. It contains 293 true negatives, 59 false positives, 44 false negatives, and 104 true positives. Thus, 44 of the 148 hateful memes are missed, while 59 of the 352 non-hateful memes are incorrectly flagged. Although both error types matter in content moderation, false negatives are particularly important because they allow hateful content to remain undetected.
\section{Conclusion}

We have presented SZED's system for ArGuard 2026 Track~A: a compact late-fusion
binary classifier for A1 (test macro-F1 = 0.7597), and a purpose-built multi-task
label-attention architecture for A2 (test macro-F1 = 0.3142 over 10 active
classes), improved via cascade calibration that exploits the task's own label
hierarchy.
Raw vs.\ calibrated scores on both dev and test confirm that calibration is
beneficial on both splits ($+$0.012-0.017 macro-F1 absolute).
Our ablation (Table~\ref{tab:a1-ablation}) shows that a more expressive
cross-modal attention mechanism and an explicit incongruity feature do not automatically help
on this dataset:
the CLIP cosine similarity between text and image carries no separating information
between hateful and non-hateful memes here, a finding consistent with
the AHA-Memes observation that caption semantics dominate image content as the
primary signal \cite{kmainasi2026ahamemes}.
For future work, we would like to explore dialect-aware pretraining or data
augmentation for the underrepresented A2 classes, and test whether the
incongruity signal becomes informative when computed with an Arabic-specialized
CLIP variant rather than a general
multilingual one.

\section*{Limitations}

We did not preserve trained model weights between the development and test phases
of the leaderboard, so each reported result comes from an independently trained
run of the same architecture.
We mitigate this by reporting run-to-run variance wherever we have multiple
test-phase runs, but cannot rule out that a single unusually good or bad run is
driving any particular number.

The A2 per-class threshold tuning relies on 500 dev examples.
Several classes have single-digit support in that split, making per-class
thresholds susceptible to overfitting; the rarest-class results should be
interpreted with this caveat in mind.

Our A2 per-class error analysis is necessarily limited for the lowest-support
test classes (\textit{Exclusion}, n=3; \textit{Slurs}, n=16), and to a lesser
extent \textit{Dehumanization} (n=21) and \textit{Inferiority} (n=23), where F1
is high-variance by construction.

The cross-modal incongruity feature was computed with a general-purpose
multilingual CLIP variant.
Dedicated unimodal (text-only and image-only) baseline runs were omitted due to
compute constraints; we instead reference the AHA-Memes baselines
\cite{kmainasi2026ahamemes} as an external reference point.
An Arabic-specialized CLIP may yield a more informative incongruity signal, but we
did not have the compute budget to test this.

Finally, our system uses no external training data or synthetic augmentation,
leaving potential headroom unexplored for the lowest-support A2 classes.

\section*{Ethical Considerations}

This work concerns automated detection of hateful content, which carries risk in
both directions: false negatives let harmful content go unflagged, while false
positives risk over-censoring legitimate speech, satire, or in-group reclaimed
language, particularly given the cultural and dialectal nuance of Arabic online
discourse.
Our A1 error analysis identifies false negatives (44 missed hateful memes out of
148) as the dominant error mode, and we note this explicitly so that downstream
users can calibrate their reliance on the system accordingly.
All training and evaluation data is the officially released, consented ArGuard
shared-task dataset; we did not collect, scrape, or introduce any additional
user-generated content.
Potentially offensive content (slurs, dehumanizing language) is handled only in
aggregate statistical form, and we did not reproduce specific offensive text or
imagery.
We would caution against deploying this system or any system trained on 3,500
examples with several classes under 60 instances as a sole or automated moderation decision-maker without human review, given its demonstrated instability on rare categories and the inherent difficulty of the task.

\section*{Acknowledgements}
We thank the ArGuard 2026 organizers for the shared task and dataset.

\bibliography{custom_}

\clearpage
\appendix
\section*{Appendix}
\section{Methodology on Different Models}
\label{sec:appendix}
 
\subsection{Subtask A1}
\label{sec:appendix-a1}
 
We tested our A1 methodology on three other Arabic text-model variants in place of
MARBERTv2, namely AraBERTv0.2 \cite{antoun-etal-2020-arabert}, CAMeLBERT-DA, and
CAMeLBERT-Mix \cite{inoue-etal-2021-interplay} (best variant per model, test set).
As Table~\ref{tab:a1-modelcomp} shows, all four models land in a narrow
0.7550-0.7610 macro-F1 band, with differences smaller than our run-to-run variance
(Section~5.1).
We read this as the fusion design and threshold tuning carrying most of the
performance, with the choice of Arabic text encoder a second-order effect for A1.
 
\begin{table}[H]
\centering
\small
\begin{tabular}{lr}
\toprule
Model Backbone & Test Macro-F1 \\
\midrule
AraBERTv0.2    & \textbf{0.7610} \\
MARBERTv2      & 0.7597 \\
CAMeLBERT-Mix  & 0.7560 \\
CAMeLBERT-DA   & 0.7550 \\
\bottomrule
\end{tabular}
\caption{A1 methodology on different backbone models (test set). All models fall
within a narrow range of one another.}
\label{tab:a1-modelcomp}
\end{table}
 
\subsection{Subtask A2}
\label{sec:appendix-a2}
 
We tested our A2 methodology on the same three
alternative Arabic models used for A1. Here the choice of model matters more: CAMeLBERT-Mix reaches 0.3296 test macro-F1, ahead of MARBERTv2 (0.3142), CAMeLBERT-DA (0.3112), and AraBERTv0.2 (0.3040), as shown in Table~\ref{tab:a2-modelcomp}. CAMeLBERT-Mix leads clearly on test, ahead of the next-best model (MARBERTv2) by 0.0154 macro-F1. Our official leaderboard submission used MARBERTv2 (0.3142, the most recent run); the gap suggests CAMeLBERT-Mix is the natural first choice for an improved A2 submission.
 
\begin{table}[H]
\centering
\small
\begin{tabular}{lr}
\toprule
Model Backbone & Test Macro-F1 \\
\midrule
CAMeLBERT-Mix  & \textbf{0.3296} \\
MARBERTv2      & 0.3142 \\
CAMeLBERT-DA   & 0.3112 \\
AraBERTv0.2    & 0.3040 \\
\bottomrule
\end{tabular}
\caption{A2 methodology on different backbone models
(test macro-F1). CAMeLBERT-Mix leads clearly on this
split. The MARBERTv2 row reflects our most recent run (the official leaderboard submission, 0.3142).}
\label{tab:a2-modelcomp}
\end{table}
 
\end{document}